\documentclass[10pt,twocolumn,letterpaper]{article}

\usepackage[pagenumbers]{cvpr} 

\usepackage{soul}
\usepackage{multirow}
\usepackage{wrapfig}
\usepackage{makecell}
\usepackage[accsupp]{axessibility}  
\newcommand{\printfnsymbol}[1]{%
  \textsuperscript{\@fnsymbol{#1}}%
}

\definecolor{cvprblue}{rgb}{0.21,0.49,0.74}
\usepackage[pagebackref,breaklinks,colorlinks,allcolors=cvprblue]{hyperref}

\def\paperID{10347} 
\def\confName{CVPR}
\def\confYear{2025}

\title{An Image-like Diffusion Method for Human-Object Interaction Detection}

\author{Xiaofei Hui \quad Haoxuan Qu\textsuperscript{†} \quad Hossein Rahmani \quad Jun Liu\\
Lancaster University\\
{\tt\small \{x.hui, h.qu5, h.rahmani, j.liu81\}@lancaster.ac.uk}
}

\begin{document}
\maketitle
\def\thefootnote{{\textsuperscript{†}}}\footnotetext{Corresponding Author}
\begin{abstract}
Human-object interaction (HOI) detection often faces high levels of ambiguity and indeterminacy, as the same interaction can appear vastly different across different human-object pairs. Additionally, the indeterminacy can be further exacerbated by issues such as occlusions and cluttered backgrounds. To handle such a challenging task, in this work, we begin with a key observation: the output of HOI detection for each human-object pair can be recast as an image. Thus, inspired by the strong image generation capabilities of image diffusion models, we propose a new framework, \textbf{HOI-IDiff}. In \textbf{HOI-IDiff}, we tackle \textbf{HOI} detection from a novel perspective, using an \textbf{I}mage-like \textbf{Diff}usion process to generate HOI detection outputs as images. Furthermore, recognizing that our recast images differ in certain properties from natural images, we enhance our framework with a customized HOI diffusion process and a slice patchification model architecture, which are specifically tailored to generate our recast ``HOI images''. Extensive experiments demonstrate the efficacy of our framework. 
\end{abstract}    
\section{Introduction}
\label{sec:intro}

Human-object interaction (HOI) detection aims to predict all interactions between each human-object pair in an image, with each interaction represented as a \textit{$<$human-interaction-object$>$} triplet. 
Its significance spans across diverse applications, such as robotics \cite{mascaro2023hoi4abot}, visual question answering \cite{chen2020counterfactual}, and image captioning \cite{yao2018exploring}. 
Recently, it has gained a great deal of research attention \cite{zhang2022efficient,zhang2023exploring,li2024disentangled,luo2024discovering,wu2024exploring,wang2024bilateral,wang2024cyclehoi,yang2024open}.
Despite growing efforts, HOI detection still remains a challenging task.
One reason is that the same interaction can look very differently across different human-object pairs, while different interactions can sometimes appear visually similar \cite{chen2024uahoi,bai2023automatically}.
Additionally, challenges also arise from other factors such as occlusions and cluttered backgrounds \cite{wang2024bilateral,kilickaya2020diagnosing,zhou2022human}.
These challenges can lead to high noise and indeterminacy (ambiguity) in the HOI detection process, resulting in error-prone predictions \cite{chen2024uahoi}. 

Recently, showing superb performance in generating high-quality images, image diffusion models \cite{ho2020denoising,peebles2023scalable} have become powerful image generators across various fields \cite{kidder2024advanced,diffusion_generation_survey}. 
Specifically, these models typically transform noisy, indeterminate images into high-quality, determinate ones by gradually removing the noise.
This process intuitively breaks down the potentially large gap between noisy images and high-quality ones into a series of smaller steps \cite{song2019generative}, making the transition smooth and quite manageable. 

\begin{figure}[t]
  \centering
   \includegraphics[width=\columnwidth]{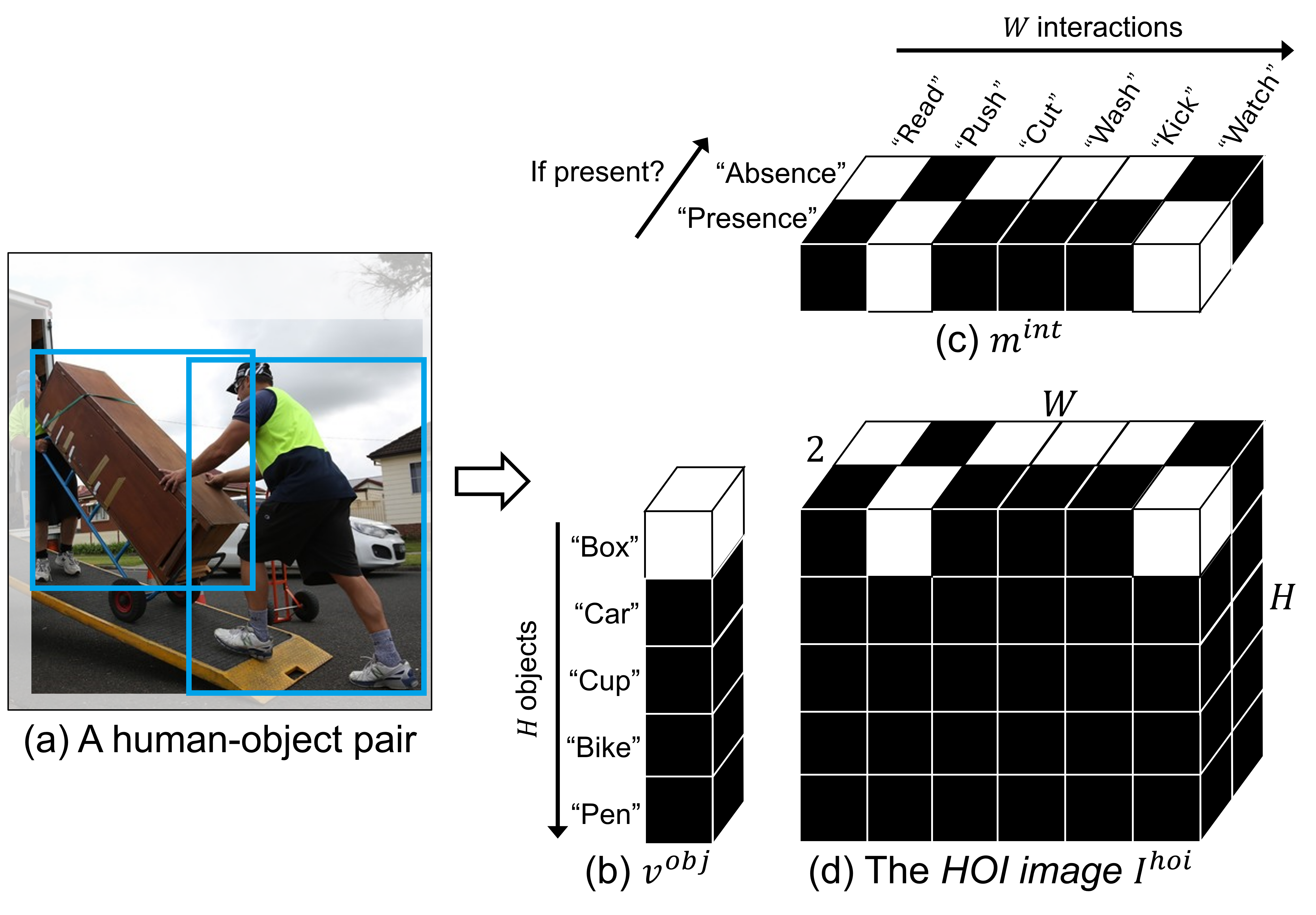}
   \vspace{-0.4cm}
   \caption{Illustration of the \textit{HOI image} (of shape $H \times W \times 2$) formed over a human-object pair in (a) in which the human is watching and pushing the box. 
   In this figure, we illustrate the case in which $H = 5$ and $W = 6$.
   We use a black pixel to represent a pixel with value 0, and a white pixel to represent a pixel with value 1. 
   As shown, the \textit{HOI image} $I^{hoi}$ in (d) is formed as the product of the vector $v^{obj}$ in (b) and the matrix $m^{int}$ in (c), in which $v^{obj}$ (of shape $H$) represents the object classification result, and $m^{int}$ (of shape $W \times 2$) represents the interaction prediction result. 
   }
   \vspace{-0.2cm}
   \label{fig:1}
\end{figure}

Motivated by the efficacy of image diffusion models, in this work, for HOI detection—a task that also involves noise and indeterminacy—we explore \textit{tackling it through an image diffusion process}. We start our exploration with a key observation: the outputs of HOI detection for each human-object pair can also \textit{form an image}. Specifically, since HOI detection for each human-object pair can be viewed as two interrelated sub-tasks, i.e., object classification and interaction prediction \cite{zheng2023open}, the output of HOI detection for each human-object pair can then be represented as an image in three steps:
\textbf{(1) Object classification.} Firstly, let $H$ denote the total number of object categories. As shown in Fig.~\ref{fig:1} (b), the output of object classification can be represented as a probability distribution $v^{obj}$ of size $H$. Especially, in the human-object pair illustrated in Fig.~\ref{fig:1}, the object category is ``box''. Thus, in $v^{obj}$, the entry corresponding to the ``box'' category has a value of 1.
\textbf{(2) Interaction prediction.} Next, for the sub-task interaction prediction, it can be treated as a set of binary classification problems \cite{gkioxari2018detecting,huynh2021interaction}, each determining the presence/absence of a certain interaction category within the current human-object pair. Consequently, let $W$ denote the total number of interaction categories. As shown in Fig.~\ref{fig:1} (c), the output of interaction prediction can then be represented as $W$ probability distributions, each of size 2, collectively denoted as $m^{int}$ of size $W \times 2$. Especially, in the human-object pair shown in Fig.~\ref{fig:1}, the interactions ``push'' and ``watch'' are present. Hence, in $m^{int}$, the two entries corresponding to the ``presence'' of ``push'' and ``watch'' both have values of 1.
\textbf{(3) Interrelation capturing.} Finally, the interrelation between the above two sub-tasks can be captured via ``multiplying'' their corresponding outputs (i.e., $v^{obj}$ and $m^{int}$). 
Specifically, to allow this multiplication, we introduce an additional dimension of size $1$ to both $v^{obj}$ and $m^{int}$ (i.e., now $v^{obj}$ is of size $H \times 1$ and $m^{int}$ is of size $1 \times W \times 2$). Then, via multiplying $v^{obj}$ and $m^{int}$, as shown in Fig.~\ref{fig:1} (d), we can form the HOI detection output per human-object pair as an image $I^{hoi}$ of shape $H \times W \times 2$, with height $H$, width $W$, and two channels. We call it an \textit{HOI image}.

Based on the above, {inspired by \cite{foo2024action}}, we observe that, we can \textit{recast HOI detection from a novel perspective as an HOI image generation problem}, and obtain accurate HOI detection outputs via generating high-quality \textit{HOI images}. To generate such high-quality \textit{HOI images}, we can then use image diffusion models \cite{ho2020denoising,peebles2023scalable} with their strong image generation abilities. To this end, in this work, we propose a novel framework \textbf{HOI-IDiff} performing \textbf{HOI} detection via an \textbf{I}mage-like \textbf{Diff}usion process, as outlined below.

Overall, following existing HOI detection methods \cite{hou2020visual,zheng2023open,yuan2023rlipv2}, our framework first uses a pre-trained detector to prepare human and object bounding boxes, then handles the two interrelated sub-tasks of HOI detection: object classification and interaction prediction. Particularly, our framework tackles these two interrelated sub-tasks via optimizing the image diffusion model to generate a high-quality \textit{HOI image} for each human-object pair. In specific, following existing diffusion model works \cite{ho2020denoising,peebles2023scalable,foo2024action}, our framework's optimization of the image diffusion model operates in two phases: the forward process and the reverse process.
In the forward process, the goal is to progressively add noise to the ground truth \textit{HOI images} (an example is shown in Fig.~\ref{fig:1}), creating supervision signals for the reverse process to train the image diffusion model. Conversely, the reverse process aims to optimize the image diffusion model to reverse the forward process, generating high-quality \textit{HOI images} from which accurate HOI detection results can be obtained.

However, we emphasize that, directly using a typical image diffusion model \cite{ho2020denoising,peebles2023scalable} to generate \textit{HOI images} in our framework may not yield optimal results.
This is because, typical image diffusion models are designed to generate natural images, whereas \textit{HOI images} differ in certain properties from natural images.
Specifically, as mentioned above, an \textit{HOI image} is created as the product of its corresponding vector $v^{obj}$ and matrix $m^{int}$, in which $v^{obj}$ is a probability distribution, and $m^{int}$ consists of $W$ probability distributions.
As a result, an \textit{HOI image} essentially also represents $W$ \textit{joint probability distributions}. 
Considering this unique property of the \textit{HOI image}, in our framework, rather than performing a typical diffusion process \cite{ho2020denoising,peebles2023scalable}, we instead introduce novel designs, specially customizing the diffusion process for generating high-quality \textit{HOI images}.

Moreover, unlike natural images in which neighboring pixels tend to be more directly correlated, in \textit{HOI images}, more direct correlations are typically found among pixels within the same row or column. 
Hence, the architectures of existing image diffusion models \cite{ho2020denoising,peebles2023scalable}, which tend to focus on local neighborhood processing to effectively handle natural images, may be ill-suited for dealing with our \textit{HOI images}.
To handle this, we further adapt the diffusion model in our framework with a slice patchification architecture, specifically optimizing it for processing \textit{HOI images}.

The contributions of our work include:
1) From a novel perspective, we formulate HOI detection as an \textit{HOI image} generation problem. We then handle this task via leveraging the strong image generation capabilities of image diffusion models to generate high-quality \textit{HOI images}.
2) We incorporate several designs in our framework to further ensure the compatibility of the image diffusion model and the \textit{HOI images}, from the perspectives of both the diffusion process and the model architecture.
3) Our framework achieves state-of-the-art performance on the evaluated benchmarks.

\section{Related Work}
\label{sec:related_work}

\noindent\textbf{HOI Detection.} Due to the wide range of applications, HOI detection has received a lot of attention \cite{fang2021dirv,kim2020uniondet,liao2020ppdm,wang2020learning,zhong2021glance,ma2023fgahoi,bansal2020detecting,li2020detailed,xu2019learning,fang2018pairwise,gupta2019no,wan2019pose,park2023viplo,qi2018learning,gao2018ican,yuan2022rlip,gkioxari2018detecting,hou2020visual,hou2021affordance,li2019transferable,zheng2023open}, and most of recent methods fall into two categories: one-stage methods and two-stage methods. Among them, one stage methods, such as UnionDet \cite{kim2020uniondet}, PPDM \cite{liao2020ppdm}, and GGNet \cite{zhong2021glance}, generally train an one-stage HOI detector to detect HOI triplets in a single forward pass.
Besides one-stage methods, two-stage methods have also attracted much attention recently, due to their training efficiency \cite{zhang2022efficient,zhang2023exploring}.

In two-stage methods, most approaches first use an off-the-shelf detector to identify all humans and objects in the input image, followed by the recognition of interactions between every human-object pair. Qi et al. \cite{qi2018learning} introduced a method that represents HOI structures as graphs, incorporating graph neural networks to enhance HOI detection. Fang et al. \cite{fang2018pairwise} leveraged correlations among different human body parts to improve the identification of human-object interactions. More recently, Zhang et al. \cite{zhang2023exploring} explored how to better utilize visual contextual cues from the input image to enhance the performance of HOI detection.

In this work, like existing two-stage methods, we also incorporate a pre-trained detector into our framework. Yet, unlike previous works, we introduce a novel perspective by reframing HOI detection as an \textit{HOI image} generation task. This allows us to leverage the strong image generation abilities of image diffusion models to perform HOI detection.

\noindent\textbf{Diffusion Models} \cite{ho2020denoising,sohl2015deep,peebles2023scalable} are originally introduced for image generation, where they show impressive capabilities.
Since then, diffusion models have also been applied to various tasks, such as object counting \cite{hui2024harnessing}, trajectory prediction \cite{gu2022stochastic}, and human mesh recovery \cite{foo2023distribution}. 
Broadly, most existing methods using diffusion models can be clustered into two main categories: those that use pre-trained diffusion models as implicit knowledge bases \cite{hui2024harnessing,peng2024harnessing,zhang2024diff,wang2024cyclehoi,yang2024open,khanislime,li2024human,li2023your,chen2024robust}, and those that train the diffusion model to use its powerful step-by-step denoising ability \cite{gong2023diffpose,xu20246d,foo2023distribution,foo2024action,gu2022stochastic,austin2021structured,hoogeboom2021argmax,han2022card}. 
Our method falls into the latter category. Specifically, given that \textit{HOI images} differ in certain properties from natural images, we customize the diffusion process of image diffusion models to better utilize their powerful denoising abilities for HOI detection. 

\section{Revisiting Image Diffusion Models}
\label{sec:revisiting}

The image diffusion model \cite{ho2020denoising,peebles2023scalable}, which is a type of probabilistic generative model, operates in two phases: the forward process and the reverse process. 
In specific, in the forward process, an original sample like a high-quality image ($d_0$) is progressively diffused into a completely random noise (typically Gaussian noise) $d_K \sim \mathcal{N}(\textbf{0}, \textbf{I})$ through a series of steps (i.e., $d_0 \rightarrow d_1 \rightarrow ... \rightarrow d_K$). 
In contrast, in the reverse process, the completely random noise $d_K$ is guided to be step-by-step denoised towards the sample $d_0$ (i.e., $d_K \rightarrow d_{K-1} \rightarrow ... \rightarrow d_0$).
Notably, both the forward and reverse processes follow a Markov chain structure.

\noindent\textbf{Forward Process.} The purpose of the forward process is to acquire necessary supervision signals, thereby enabling the training of the diffusion model in the reverse process.
To achieve this, as the denoising in the reverse process is performed in a step-wise manner, we here first obtain both $d_K$ and the intermediate step results $\{d_k\}^{K-1}_{k=1}$ between $d_0$ and $d_K$. In specific, leveraging $d_0$, $\{d_k\}^{K}_{k=1}$ can be obtained one by one from $d_1$ to $d_{K}$ via the following equation as:
\begin{equation} \label{eq:revisiting_1}
\setlength{\abovedisplayskip}{3pt}
\setlength{\belowdisplayskip}{3pt}
\begin{aligned}
d_k = \sqrt{1-\beta_k}d_{k-1} + \sqrt{\beta_k}\epsilon
\end{aligned}
\end{equation}
where $\{\beta_k \in (0, 1)\}^K_{k=1}$ is a set of controllers that regulate the scale of the injected noise at each step, and $\epsilon \sim \mathcal{N}(\textbf{0}, \textbf{I})$. 
With the help of Eq.~\ref{eq:revisiting_1}, following \cite{ho2020denoising}, we can further formulate $d_k$ directly from $d_0$ in a single step as:
\begin{equation} \label{eq:revisiting_2}
\setlength{\abovedisplayskip}{3pt}
\setlength{\belowdisplayskip}{3pt}
\begin{aligned}
d_k = \sqrt{\overline{\alpha}_k}d_0 + \sqrt{1 - \overline{\alpha}_k}\epsilon
\end{aligned}
\end{equation}
where $\alpha_k = 1 - \beta_k$ and $\overline{\alpha}_k = \prod^k_{s=1} \alpha_s$. According to Eq.~\ref{eq:revisiting_1} and Eq.~\ref{eq:revisiting_2}, following \cite{ho2020denoising}, we can finally derive the supervision signal $q(d_{k-1}|d_k, d_0)$ that will be used for training during the reverse process as:
\begin{equation} \label{eq:revisiting_4}
\setlength{\abovedisplayskip}{3pt}
\setlength{\belowdisplayskip}{3pt}
\begin{aligned}
q(d_{k-1}|d_k, d_0) = \mathcal{N}(d_{k-1}; \widetilde{\mu}_k(d_k, d_0), \widetilde{\beta}_k\textbf{I})
\end{aligned}
\end{equation}
where $\widetilde{\mu}_k(d_k, d_0) = \frac{\sqrt{\overline{\alpha}_{k-1}}\beta_k}{1-\overline{\alpha}_k} d_0 + \frac{\sqrt{\alpha_{k}}(1 - \overline{\alpha}_{k-1})}{1-\overline{\alpha}_k} d_k$ and $\widetilde{\beta}_k = \frac{1 - \overline{\alpha}_{k-1}}{1 - \overline{\alpha}_k}\beta_k$.

\noindent\textbf{Reverse Process.} Given $q(d_{k-1}|d_k, d_0)$, in the reverse process, the diffusion model $\theta$ is trained to predict the conditional probability $p_{\theta}(d_{k-1}|d_k)$, utilizing $q(d_{k-1}|d_k, d_0)$ as the supervision signal. Leveraging $\theta$, we can formulate each reverse diffusion step as a function $g$ with $d_k$ and $\theta$ as its inputs and with $d_{k-1}$ as its output, i.e., $d_{k-1} = g(d_k, \theta)$. Note that, once the diffusion model has been trained, the forward process is no longer required. Instead, during inference, we only perform the reverse process, in which the trained diffusion model is used to convert random Gaussian noise $d_K \sim \mathcal{N}(\textbf{0}, \textbf{I})$ into a sample $d_0$ of the target distribution.

\section{Proposed Method}
\label{sec:method}

In this work, from a novel perspective, we propose to reformulate HOI detection as an \textit{HOI image} generation problem (as elaborated in Sec.~\ref{sec:recast}), and tackle this task via utilizing the image diffusion model to generate high-quality \textit{HOI images}. 
However, because \textit{HOI images} differ from natural images, naively applying the typical image diffusion models \cite{ho2020denoising,peebles2023scalable} to generate \textit{HOI images} can be sub-optimal.
Considering this, we here propose a novel HOI-IDiff framework, which is specially designed for the generation of \textit{HOI images}. 
Specifically, in HOI-IDiff, we incorporate the diffusion model with a new \textit{HOI image} diffusion process (as discussed in Sec.~\ref{sec:diffusion}), as well as a new slice patchification model architecture (as introduced in Sec.~\ref{sec:architecture}). 

\subsection{Problem Reformulation}
\label{sec:recast}

Below, we show how we reformulate HOI detection as an image generation problem, taking a pair of human and object as an example. 
Previous works \cite{zheng2023open,hou2020visual} have shown that, HOI detection for each human-object pair can be viewed as two interrelated sub-tasks: object classification and interaction prediction.
Hence, in this work, we handle these two sub-tasks jointly. Specifically, as shown in Fig.~\ref{fig:1}, the HOI detection output for each human-object pair can be formulated as the product of a vector $v^{obj}$ and a matrix $m^{int}$, which respectively denote the outputs of the object classification and interaction prediction sub-tasks.

Here, $v^{obj}$ of size $H$ represents a discrete probability distribution, where $H$ is the total number of object categories. In specific, in $v^{obj}$, the $h$-th element $v^{obj}[h]$ represents the probability that the object in the current human-object pair belongs to the $h$-th object category. Meanwhile, $m^{int}$ of size $W \times 2$ represents a collection of $W$ discrete probability distributions, each of size 2, where $W$ is the total number of interaction categories. Specifically, in the $w$-th distribution of $m^{int}$ (i.e., $m^{int}[w]$), the first element $m^{int}[w, 0]$ and the second element $m^{int}[w, 1]$ respectively represent the presence and absence probabilities of the $w$-th interaction in the current human-object pair.

With $v^{obj}$ and $m^{int}$ defined as above, via multiplying them, we can represent the HOI detection output for each human-object pair as an \textit{HOI image} $I^{hoi}$ of size $H \times W \times 2$, with height $H$, width $W$, and two channels, {as shown in Fig.~\ref{fig:1}.} 
Notably, in $I^{hoi}$, the $w$-th vertical slice $I^{hoi}[:, w, :]$ of size $H \times 2$ is derived as the product of two distributions (i.e., $v^{obj}$ of size $H$ and $m^{int}[w]$ of size $2$). This leads the \textit{HOI image} to hold a unique property: each of its vertical slice essentially represents a joint probability distribution that sums to 1. 
Note that, this property also holds conversely. Specifically, for a matrix of size $H \times W \times 2$, as long as each of its vertical slice forms a probability distribution summing to one, it represents an \textit{HOI image}, and we can uniquely decompose it into its corresponding vector $v^{obj}$ and matrix $m^{int}$ (with the analysis provided in supplementary).

Observing that the outputs of HOI detection can be reformulated as two-channel \textit{HOI images} via above, we thus are motivated to reframe HOI detection as an image generation problem, harnessing the powerful image generation capabilities of image diffusion models.

\subsection{HOI Image Diffusion Process}
\label{sec:diffusion}

From the above, we observe that, generating high-quality \textit{HOI images} enables us to achieve accurate HOI detection outputs. 
This allows us to recast HOI detection as an \textit{HOI image} generation problem, and tackle HOI detection by focusing on generating high-quality \textit{HOI images}. In this section, we then introduce how we generate high-quality \textit{HOI images} in our framework, via an image diffusion process.
In specific, though typical (natural) image diffusion models \cite{ho2020denoising,peebles2023scalable} could be naively applied here to generate \textit{HOI images}, as \textit{HOI images} differ in certain properties from natural images, this naive application can lead to sub-optimal performance. 
Hence, in this work, we aim to propose a novel \textit{HOI image} diffusion process tailored specifically for \textit{HOI image} generation.
Yet, designing such a diffusion process can be challenging. This is because, as mentioned below, to make the designed diffusion process suitable for the HOI detection task and for the \textit{HOI images}, the process is expected to hold the following three attributes:

\ul{Attribute 1:} Firstly, recall that like existing HOI detection works \cite{zhang2022efficient,zhang2023exploring,hou2020visual,zheng2023open}, our framework begins by passing the input image through a pre-trained detector. In this step, besides the human and object bounding boxes, the detector also provides a prior categorical distribution for each detected object, in the form of a probability vector of length $H$. These distributions encode valuable prior information about object categories. Thus, in our framework, {rather than starting denoising from completely random noise, for each identified human-object pair, we would like to first initialize a noisy \textit{HOI image} using the prior categorical distribution of the object in the pair}. 
Next, we aim to {instead use this noisy (yet not completely random) \textit{HOI image} as the starting point for the denoising process in the reverse diffusion process}. 
This allows us to effectively leverage prior information, guiding the diffusion model to generate more accurate \textit{HOI images}. Notably, to achieve the above and satisfy this attribute, during the forward diffusion process, \textbf{we need to diffuse towards the initialized noisy \textit{HOI image} instead of towards completely random noise}, for acquiring appropriate supervision signals for the training of the diffusion model.

\begin{figure*}[t]
\centering
\includegraphics[width=0.89\textwidth]{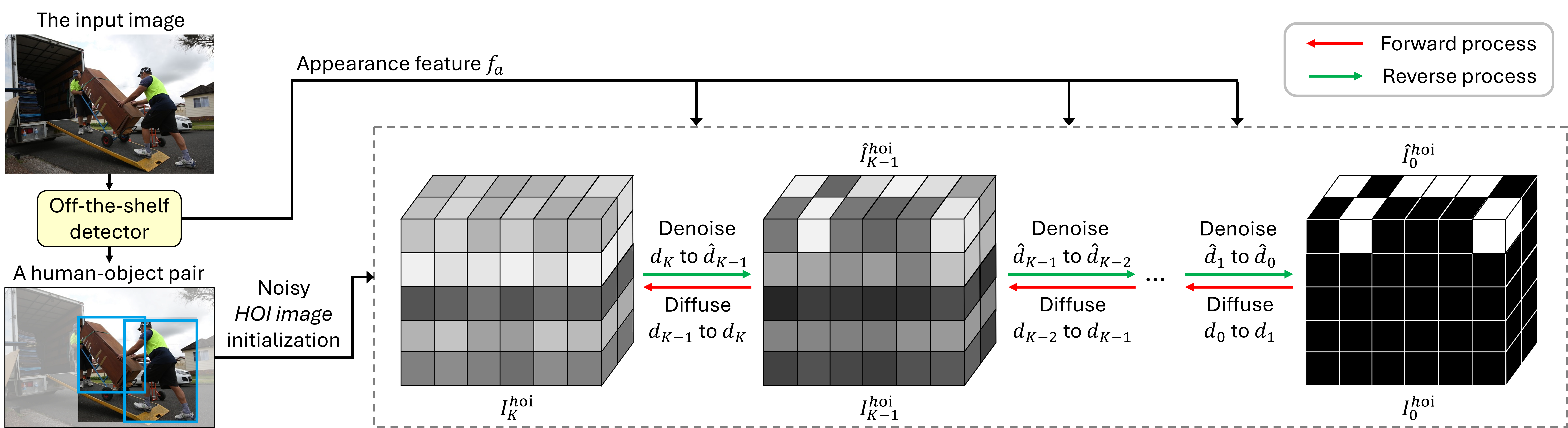}
\vspace{-0.2cm}
\caption{Illustration of our \textit{HOI image} diffusion process. As indicated by the red arrows from right to left, the forward \textit{HOI image} diffusion process gradually diffuses the ground-truth \textit{HOI image} $I^{hoi}_0$ towards $I^{hoi}_K$ (i.e., $I^{hoi}_0 \rightarrow ... \rightarrow I^{hoi}_{K-1} \rightarrow I^{hoi}_{K}$). Conversely, as shown by the green arrows from left to right, in the reverse \textit{HOI image} diffusion process, conditioned on the appearance feature $f_a$, the diffusion model $\theta$ is guided to progressively reconstruct a desired high-quality \textit{HOI image} $\hat{I}^{hoi}_0$ from $I^{hoi}_K$ (i.e., $I^{hoi}_K \rightarrow \hat{I}^{hoi}_{K-1} \rightarrow ... \rightarrow \hat{I}^{hoi}_0$).}
\vspace{-0.3cm}
\label{fig:2}
\end{figure*}

\ul{Attribute 2:} Meanwhile, as explained in Sec.~\ref{sec:recast}, compared to natural images, the \textit{HOI image} possesses a unique property. That is, each of its vertical slice (of size $H \times 2$), derived as the product of two distributions (i.e., the distribution $v^{obj}$ and a distribution in $m^{int}$), essentially represents a joint probability distribution that sums to 1. Considering this, to guide the diffusion model to generate accurate \textit{HOI images} in the reverse diffusion process from noisy initializations, in the forward process, we aim to \textbf{ensure all intermediate results at different diffusion steps to retain their vertical slices as probability distributions that sum to 1}. 
As mentioned in Sec.~\ref{sec:recast}, this ensures that the intermediate result at each diffusion step essentially forms an \textit{HOI image}.

\ul{Attribute 3:} With the above two attributes in place, the proposed diffusion process shall be made compatible for the \textit{HOI image} generation problem. 
Meanwhile, we observe that, in existing diffusion works \cite{ho2020denoising}, the formulation of the supervision signal $q(d_{k-1}|d_k, d_0)$ in the forward diffusion process is a crucial step \cite{foo2024action}. 
Specifically, with $q(d_{k-1}|d_k, d_0)$ formulated, during each reverse step, once the probability distribution $p_{\theta}(d_{k-1}|d_k)$ is predicted from the diffusion model, the difference between $q(d_{k-1}|d_k, d_0)$ and $p_{\theta}(d_{k-1}|d_k)$ can be then used to optimize the diffusion model effectively \cite{ho2020denoising,foo2024action}.
Hence, in our diffusion process, we aim to still \textbf{allow for the formulation of $q(d_{k-1}|d_k, d_0)$}.

Above, we introduce the three attributes that an image diffusion process is urged to satisfy for it to generate high-quality \textit{HOI images}. Notably, based on these attributes, a typical natural image diffusion model is clearly not suitable for generating \textit{HOI images}, as it satisfies neither \ul{Attribute 1} nor \ul{Attribute 2}. 
In light of this, to generate high-quality \textit{HOI images}, in this work, we propose a novel \textit{HOI image} diffusion process that comprehensively satisfies all the above three attributes. 
Below, we first describe how our proposed process initializes the noisy \textit{HOI image} mentioned in \ul{Attribute 1}. 
We then discuss the corresponding forward and reverse processes in our proposed approach.
We also illustrate the \textit{HOI image} diffusion process in Fig.~\ref{fig:2}.

\textbf{Noisy \textit{HOI image} initialization.} To initialize the noisy \textit{HOI image} for a certain human-object pair such that it can be used as the start point of the reverse diffusion process, as described in Sec.~\ref{sec:recast}, we simply need to initialize its corresponding vector  $v^{obj}$ and matrix $m^{int}$.
Among which, for $v^{obj}$, as mentioned above, a prior distribution has been provided by an off-the-shelf detector in the form of a probability vector, so we directly set $v^{obj}$ to be this vector.
Yet for $m^{int}$, since no prior is provided by the detector, we here simply set all values in $m^{int}$ to 0.5, initializing each interaction category with an equal probability of being present or absent in the current human-object pair.
Via the above, we can derive both $m^{int}$ and $v^{obj}$. Their product can then be used to initialize the noisy \textit{HOI image} accordingly.

\textbf{Forward process.} After introducing the initialization of the noisy \textit{HOI image}, we now explain the forward process of our \textit{HOI image} diffusion method. For simplicity, we illustrate this process by focusing on a single vertical slice of the \textit{HOI image}. Let $d_0$ represent the $w$-th vertical slice of the ground-truth \textit{HOI image} $I^{hoi}_0$, and let $K$ denote the total number of diffusion steps. The forward process begins by diffusing $d_0$ to derive $\{d_k\}_{k=1}^K$ (i.e., $d_0 \rightarrow d_1 \rightarrow ... \rightarrow d_K$). We then use $\{d_k\}_{k=0}^K$ to obtain the posterior distributions $\{q(d_{k-1}|d_k, d_0)\}_{k=1}^K$, which serve as supervision signals for training the diffusion model in the reverse process.

In particular, during deriving $d_k$, to ensure that our \textit{HOI image} diffusion process satisfies the above-mentioned attributes, let $d_{init}$ denote the $w$-th vertical slice of the noisy \textit{HOI image} initialized in the current human-object pair. Our initial aim is to guide $d_k$ to gradually converge towards $d_{init}$ as $k$ becomes large. To achieve this, we get inspired by that given suitable input parameters, a Multinomial distribution $P^{Mu}$ can converge to any discrete probability distributions (e.g., $d_{init}$) \cite{Soch2024-bz}. Motivated by this, we perform our forward \textit{HOI image} diffusion process as follows.

Specifically, at each diffusion step, to acquire $d_k$ from $d_{k-1}$, similar to the typical natural image forward diffusion process in Eq.~\ref{eq:revisiting_1}, we inject a small noise to $d_{k-1}$ as:
\begin{equation} \label{eq:forward_1}
\setlength{\abovedisplayskip}{3pt}
\setlength{\belowdisplayskip}{3pt}
\begin{aligned}
d_k = (1-\beta_k)d_{k-1} + \beta_k\epsilon^{Mu}
\end{aligned}
\end{equation}
However, in Eq.~\ref{eq:forward_1}, we make two changes compared to Eq.~\ref{eq:revisiting_1}. (1) Firstly, instead of from $\mathcal{N}(\textbf{0}, \textbf{I})$, we sample $\epsilon^{Mu}$ from the Multinomial distribution $P^{Mu}_T(T, d_{init})$, where $T$ is a hyperparameter representing the Multinomial distribution's number of trials, and $d_{init}$ provides the sampling probability for the distribution in each trial. Moreover, the subscript $T$ means that, after finishing the $T$ sampling trials, we further divide the sampling result by $T$ to ensure that elements in $\epsilon^{Mu}$ sum to 1.
By sampling $\epsilon^{Mu} \sim P^{Mu}_T(T, d_{init})$, we would later show that this can enable $d_k$ to finally converge to $d_{init}$ as $k$ becomes large. (2) Secondly, instead of $\sqrt{1-\beta_k}$ and $\sqrt{\beta_k}$ in Eq.~\ref{eq:revisiting_1}, $1-\beta_k$ and $\beta_k$ is used to respectively regulate $d_{k-1}$ and $\epsilon^{Mu}$ in Eq.~\ref{eq:forward_1}. By doing so, as $(1-\beta_k) + \beta_k = 1$, for $d_{k-1}$ and $\epsilon^{Mu}$ as two probability distributions each sums to 1, we can ensure $d_k$ at each diffusion step $k$ to also maintain a probability distribution sums to 1, satisfying \ul{Attribute 2}.

Next, we show that, via mapping $d_{k-1}$ to $d_k$ via Eq.~\ref{eq:forward_1}, our \textit{HOI image} diffusion process can also satisfy \ul{Attribute 1}. Specifically, with the help of Eq.~\ref{eq:forward_1}, $d_k$ can be derived from $d_0$ approximately as the following (with the derivation provided in supplementary):
\begin{equation} \label{eq:forward_2}
\setlength{\abovedisplayskip}{3pt}
\setlength{\belowdisplayskip}{3pt}
\begin{aligned}
d_k = \overline{\alpha}_k d_0 + (1 - \overline{\alpha}_k) \overline{\epsilon}^{Mu}
\end{aligned}
\end{equation}
where $\alpha_k = 1 - \beta_k$, $\overline{\alpha}_k = \prod^k_{s=1} \alpha_s$, $\overline{\epsilon}^{Mu} \sim P^{Mu}_{S_kT}(S_kT, d_{init})$, and $S_k = \frac{(1-\overline{\alpha}_k)^2}{(\prod_{j=2}^k \alpha_k)^2 \beta_1^2 + (\prod_{j=3}^k \alpha_k)^2 \beta_2^2 + ... + \beta_k^2}$. Next, from Eq.~\ref{eq:forward_2}, we can observe that, when $K$ becomes large and $\overline{\alpha}_K$ correspondingly decreases to nearly zero, the distribution of $d_K$ reaches the Multinomial distribution $P^{Mu}_{S_kT}(S_kT, d_{init})$ (i.e., $d_K \sim P^{Mu}_{S_kT}(S_kT, d_{init})$). In this case, leveraging the property of the Multinomial distribution \cite{Soch2024-bz,foo2024action}, we have $\mathbb{E}[d_K] = d_{init}$. In other words, from Eq.~\ref{eq:forward_2}, we can conclude that our forward diffusion process does gradually diffuse towards the initialized noisy \textit{HOI image}. This ensures that our diffusion process also satisfies \ul{Attribute 1}.

At this point, our \textit{HOI image} diffusion process has satisfied \ul{Attribute 1} and \ul{Attribute 2}. Next, at the final stage of our forward process, we would like to satisfy \ul{Attribute 3}, and obtain the supervision signal $q(d_{k-1}|d_k, d_0)$. Specifically, based on the properties of the Markov chain structure \cite{foo2024action}, $q(d_{k-1}|d_k, d_0)$ in our framework can be derived as:
\begin{equation} \label{eq:forward_3}
\setlength{\abovedisplayskip}{3pt}
\setlength{\belowdisplayskip}{3pt}
\begin{aligned}
& q(d_{k-1}|d_k, d_0)  = \Big(\gamma_k \big(P^{Mu}_{\frac{T(d_k - (1-\beta_k)d_{k-1})}{\beta_k}}(d_{k-1}; T, d_{init}, d_k)\big) \\
& \times \big(P^{Mu}_{\frac{S_{k-1}T(d_{k-1} - \overline{\alpha}_{k-1}d_0)}{1 - \overline{\alpha}_{k-1}}}(d_{k-1}; S_{k-1}T, d_{init}, d_0)\big)\Big)
\end{aligned}
\end{equation}
where $\gamma_k = \bigg( \sum_{d_{k-1}} \Big(\big(P^{Mu}_{\frac{T(d_k - (1-\beta_k)d_{k-1})}{\beta_k}}(d_{k-1}; T, d_{init}, d_k)\big)$ $\times \big(P^{Mu}_{\frac{S_{k-1}T(d_{k-1} - \overline{\alpha}_{k-1}d_0)}{1 - \overline{\alpha}_{k-1}}}(d_{k-1}; S_{k-1}T, d_{init}, d_0)\big)\Big)\bigg)^{-1}$. We also include more details w.r.t. the derivation of Eq.~\ref{eq:forward_3} in Supplementary. Furthermore, the formulation of $\gamma_k$ also reveals that it does not vary with any observed $d_{k-1}$. Thus, during implementation, for computational efficiency, we regard $\gamma_k$ to be a fixed hyperparameter and set $\gamma_k = 1$.

\textbf{Reverse process.} In the reverse process, our objective is to guide the diffusion model $\theta$ to learn to reconstruct the desired high-quality \textit{HOI images} from their initial noisy version. For simplicity, same as in the forward process above, we here demonstrate the reverse diffusion process also by focusing on a single vertical slice of the \textit{HOI image}. The reverse process can then be formulated as a step-wise denoising procedure from $d_K$ to $\hat{d}_0$ (i.e., $d_K \rightarrow \hat{d}_{K-1} \rightarrow ... \rightarrow \hat{d}_0$). Here, we use $\hat{d}_k$ to represent the version of $d_k$ that is estimated by the diffusion model $\theta$, for distinguishing it with $d_k$ in the above forward process.

However, generating the desired $\hat{d}_0$ purely through the reverse process from $d_K$ can be challenging. This is due to the absence of essential appearance details—like those of the human and object in the current human-object pair—in $d_K$. Notably, these details can be crucial for precise HOI predictions. To address this, during our reverse process, we also incorporate these appearance details from the input image to guide the diffusion model $\theta$, enabling it to produce $\hat{d}_0$ with more accuracy. In particular, for each human-object pair, we reuse the feature extracted from the pre-trained detector w.r.t. the pair in the input image as the appearance feature $f_a$. This feature $f_a$ is then fed into the diffusion model $\theta$ to guide its reverse process.
In addition to $f_a$, following existing diffusion works \cite{ho2020denoising, song2020denoising}, at each reverse diffusion step, we also build a unique step embedding $f^k_s$ to incorporate the step number ($k$) information into the diffusion model $\theta$. This aids $\theta$ in learning to perform denoising at each reverse step. 
Via the above, we can finally form each reverse diffusion step as a function $g$, with $\hat{d}_k$, $\theta$, $f_a$, and $f^k_s$ as its inputs and with $\hat{d}_{k-1}$ as its output:
\begin{equation} \label{eq:reverse_1}
\setlength{\abovedisplayskip}{3pt}
\setlength{\belowdisplayskip}{3pt}
\begin{aligned}
\hat{d}_{k-1} = g(\hat{d}_k, \theta, f_a, f^k_s)
\end{aligned}
\end{equation}

Above, for simplicity, we introduce our \textit{HOI image} diffusion process by focusing on a single vertical slice of the \textit{HOI image}. Notably, the same diffusion process is actually applied simultaneously across all vertical slices of the \textit{HOI image}. This means, the above diffusion process is performed on the entire \textit{HOI image} as a whole.

\subsection{Model Architecture}
\label{sec:architecture}

In this section, we describe the architecture of our diffusion model. Notably, though our model could directly adopt the architecture of typical natural image diffusion models \cite{ho2020denoising,peebles2023scalable} without making any modification, doing so may not yield optimal results. This is because, in natural images, more direct correlations often exist among neighboring pixels, leading typical image diffusion models to incorporate local neighborhood processing (e.g., local convolution \cite{ho2020denoising} or local patchification \cite{peebles2023scalable}) in their architectures. Differently, \textit{HOI images} exhibit more direct correlations primarily within horizontal or vertical slices (rows or columns) of pixels. Specifically, in an \textit{HOI image}, each horizontal slice of pixels represents the interaction prediction output under a given object category, while each vertical slice of pixels demonstrates the object classification output under a certain interaction category. For instance, in the \textit{HOI image} illustrated in Fig.~\ref{fig:1}, the first horizontal slice represents the interaction prediction output when the object category is ``box''. The above implies that, in an \textit{HOI image}, instead of neighboring pixels, pixels within the same (horizontal or vertical) slice generally tend to exhibit more direct correlations. 

Considering this property of the \textit{HOI image}, in our framework, for our diffusion model to adequately process \textit{HOI images}, we design it with a \textbf{slice patchification architecture}. 
Overall, our architecture builds on DiT \cite{peebles2023scalable}, a transformer-based diffusion model architecture that has been popularly used in processing and generating natural images. 
However, unlike DiT which transforms the input image into local patches, our architecture employs a slice patchification operation.
In this operation, rather than local patches, an \textit{HOI image} of shape $H \times W \times 2$ is instead converted into $H$ horizontal slice patches (each of shape $W \times 2$) and $W$ vertical slice patches (each of shape $H \times 2$).
By doing so, this slice patchification operation encourages the diffusion model to prioritize correlations within each slice, enhancing its ability to process \textit{HOI images}.
After slice patchification, we obtain a total of $H + W$ patches. 
Next, to further capture the potential (indirect) correlations across different slices in the \textit{HOI image}, similar to \cite{peebles2023scalable}, we further convert these $H + W$ patches into $H + W$ equal-dimensional tokens, and pass the resulting tokens into a Multi-Head Self-Attention layer. 
Specifically, we perform the above patch-to-token conversion using two MLP layers.

Based on such as a slice patchification architecture, our diffusion model can then process \textit{HOI images} effectively.

\subsection{Overall Training and Testing}
\label{sec:overall}

\noindent\textbf{Training.} In our framework, during training, we first build the diffusion model with the architecture introduced in Sec.~\ref{sec:architecture}. Next, we optimize the model via the \textit{HOI image} diffusion process explained in Sec.~\ref{sec:diffusion}, and guide the model to generate high-quality \textit{HOI images} (as illustrated in Fig.~\ref{fig:2}). Specifically, during the optimization of the diffusion model, in the forward process, for each human-object pair, to ensure $I^{hoi}_K$ better represents the initialized noisy \textit{HOI image}, we diffuse $M$ samples of $I^{hoi}_K$ from the ground-truth \textit{HOI image} $I^{hoi}_0$. We then apply the reverse process to all $M$ samples, rather than relying on only a single sample of $I^{hoi}_K$. Additionally, we incorporate the MSE loss, calculated between the supervision signals and the diffusion model’s predictions at each diffusion step.

\textbf{Testing.} During testing, given an input image, similar to existing methods \cite{zhang2022efficient,hou2020visual,zheng2023open}, we first use a pre-trained detector to detect human and object bounding boxes and form human-object pairs. 
Then, following Sec.~\ref{sec:diffusion}, we initialize a set of noisy \textit{HOI images}, one for each human-object pair, and feed them parallelly into the diffusion model. 
Finally, we post-process the \textit{HOI images} predicted by the diffusion model to acquire the HOI detection results as follows.

To determine object categories, we first observe that, each object in the input image could belong to multiple human-object pairs. 
In light of this, we categorize an object by examining all human-object pairs that include it. 
Specifically, let $\hat{I}^{hoi}_{avg}$ (with shape $H \times W \times 2$) represent the average of all predicted \textit{HOI images} that have their corresponding human-object pairs to include a certain object. 
To classify this object, we first sum $\hat{I}^{hoi}_{avg}$ along its last two axes, reducing it to a vector of length $H$. We then apply the argmax operation to this vector. 
Next, to determine the interaction categories present within each human-object pair, let $\hat{I}^{hoi}_0$ (of shape $H \times W \times 2$) represent the predicted \textit{HOI image} corresponding to a specific human-object pair, and let $h$ denote the index of the above-determined object category in this pair. 
Then, we can see if the presence probability of this interaction category is larger than the absence probability (i.e., via checking if $\hat{I}^{hoi}_0[h, w, 0] > \hat{I}^{hoi}_0[h, w, 1]$) to find out if the $w$-th interaction category is present in the current pair.
During testing, we can obtain the HOI detection results from the \textit{HOI images} predicted by the diffusion model.

\section{Experiments}
\label{sec:experiments}

To evaluate the efficacy of our framework, we conduct experiments on the HICO-DET and V-COCO datasets. 

\noindent\textbf{HICO-DET} \cite{chao2018learning} is a large-scale HOI detection dataset. It contains 47,776 images, 80 object categories, and 117 interaction categories. Moreover, it consists of three subsets: (1) the full subset, including all 600 HOI triplets, (2) the rare subset, containing 138 HOI triplets with fewer than 10 training samples, and (3) the non-rare subset, comprising the remaining 462 HOI triplets. Following \cite{zhang2022efficient,zhang2023exploring,yuan2023rlipv2}, we evaluate our method under both the Default and Known object settings. 
We report the mAP metric across all subsets.

\noindent\textbf{V-COCO} \cite{gupta2015visual}, as a sub-set of MS-COCO \cite{lin2014microsoft}, is another popularly used HOI detection dataset. It contains 10,346 images, 80 object categories, and 24 interaction categories. Following \cite{zhang2022efficient,zhang2023exploring,yuan2023rlipv2}, we report the mAP metric and evaluate our method under the $AP^{S1}_{role}$ and $AP^{S2}_{role}$ settings.

\begin{table}[t]\small
\centering
\resizebox{\columnwidth}{!}{
\begin{tabular}{c|cccccc|cc}
\toprule
\multirow{3}{*}{\textbf{Method}} & \multicolumn{6}{c|}{\textbf{HICO-DET}} & \multicolumn{2}{c}{\textbf{V-COCO}} \\ [4pt]
& \multicolumn{3}{c}{Default setting} & \multicolumn{3}{c|}{Known objects setting} & & \\ 
\cline{2-4}\cline{5-7}\cline{8-9} \\ [-8pt]
& Full & Rare & Non-rare & Full & Rare & Non-rare & AP$_{role}^{S1}$ & AP$_{role}^{S2}$ \\
\midrule
Gkioxari et al. \cite{gkioxari2018detecting} & 9.94 & 7.16 & 10.77 & - & - & - & 40.0 & 48.0 \\
Kim et al. \cite{kim2020uniondet} & 17.58 & 11.72 & 19.33 & 19.76 & 14.68 & 21.27 & 47.5 & 56.2 \\
Tamura et al. \cite{tamura2021qpic} & 29.90 & 23.92 & 31.69 & 32.38 & 26.06 & 34.27 & 58.8 & 61.0 \\
Zhang et al. \cite{zhang2022efficient} & 32.62 & 28.62 & 33.81 & 36.08 & 31.60 & 37.47 & 61.3 & 67.1 \\
Liao et al. \cite{liao2022gen} & 34.95 & 31.50 & 36.08 & 38.22 & 34.94 & 39.37 & 63.6 & 65.9\\
Ma et al. \cite{ma2023fgahoi} & 37.18 & 30.71 & 39.11 & 38.93 & 31.93 & 41.02 & - & - \\
Zhang et al. \cite{zhang2023exploring} & 44.32 & 44.61 & 44.24 & 47.81 & 48.38 & 47.64 & 64.1 & 70.2 \\
Yuan et al. \cite{yuan2023rlipv2} & 45.09 & 43.23 & 45.64 & - & - & - & 72.1 & 74.1 \\
Chen et al. \cite{chen2024uahoi} & 35.78 & 29.80 & 37.56 & 37.59 & 31.36 & 39.36 & - & - \\
Wang et al. \cite{wang2024bilateral} & 39.34 & 39.90 & 39.17 & 42.24 & 42.86 & 42.05 & 65.8 & 69.9 \\
Yang et al. \cite{yang2024open} & 44.53 & 44.48 & 44.55 & - & - & - & 66.2 & 67.6 \\
Luo et al. \cite{luo2024discovering} & 45.04 & 45.61 & 44.88 & 48.16 & 48.37 & 48.09 & 71.1 & 75.6 \\
Wu et al. \cite{wu2024exploring} & 46.01 & 46.74 & 45.80 & 49.50 & 50.59 & 49.18 & 63.0 & 68.7 \\
\midrule
Ours & \textbf{47.71} & \textbf{48.36} & \textbf{47.52} & \textbf{50.56} & \textbf{51.95} & \textbf{50.14} & \textbf{73.4} & \textbf{76.1} \\ 
\bottomrule
\end{tabular}
}
\vspace{-0.2cm}
\caption{Performance comparison on the HICO-DET and V-COCO datasets.}
\label{tab:results}
\end{table}

\noindent\textbf{Implementation details.} We conduct our experiments on an RTX 6000 Ada GPU. 
We follow \cite{yuan2023rlipv2} to use DDETR \cite{zhu2021deformable} and detect objects in our framework.
During training, we optimize the diffusion model using the AdamW optimizer \cite{loshchilov2017decoupled} with an initial learning rate of 1e-4. 
We set the batch size to be 8.
Moreover, we set the hyperparameters $K$ to 50, $T$ to 2000, and $M$ to 10.
More details are in supplementary.  

\subsection{Experimental Results}

In Tab.~\ref{tab:results}, we compare our approach with existing HOI detections methods. Compared to existing methods, our framework consistently achieves the best performance across all the evaluation settings, showing its effectiveness.

\subsection{Ablation Studies}

Following previous works \cite{zhang2022efficient,wang2024bilateral}, we conduct extensive ablation experiments on the Default setting of HICO-DET. 

\begin{table}[h]
\centering
\resizebox{\columnwidth}{!}
{
\small
\begin{tabular}{l|ccc}
\hline
Method & Full & Rare & Non-rare\\
\hline
\uppercase\expandafter{\romannumeral1}: Typical natural image diffusion process & 42.50 & 42.71 & 42.44 \\
\uppercase\expandafter{\romannumeral2}: HOI-IDiff w/o initializing noisy HOI image & 44.16 & 44.60 & 44.03 \\
\uppercase\expandafter{\romannumeral3}: HOI-IDiff with noisy HOI image as condition & 45.69 & 46.34 & 45.50 \\
\uppercase\expandafter{\romannumeral4}: HOI-IDiff w/o maintaining to be distributions & 45.91 & 46.01 & 45.88 \\
\hline
\uppercase\expandafter{\romannumeral5}: HOI-IDiff with architecture in \cite{ho2020denoising} & 43.95 & 44.45 & 43.80 \\
\uppercase\expandafter{\romannumeral6}: HOI-IDiff with architecture in \cite{peebles2023scalable} & 44.16 & 44.56 & 44.04 \\
\uppercase\expandafter{\romannumeral7}: HOI-IDiff w/o horizontal slice patches & 45.11 & 45.41 & 45.02 \\
\uppercase\expandafter{\romannumeral8}: HOI-IDiff w/o vertical slice patches & 45.81 & 45.93 & 45.77 \\
\hline
HOI-IDiff & 47.71 & 48.36 & 47.52 \\
\hline
\end{tabular}}
\vspace{-0.2cm}
\caption{Evaluation on the key components of HOI-IDiff.}
\label{Tab:ablation_study_1}
\end{table}

\noindent\textbf{Impact of the key components of our HOI-IDiff framework.} In our HOI-IDiff framework, we first propose an \textit{HOI image} diffusion process to enhance the compatibility of the image diffusion model and the \textit{HOI images}. To evaluate the  efficacy of this process, we test the following four variants. In variant \uppercase\expandafter{\romannumeral1} (\textit{typical natural image diffusion process}), we directly use the typical natural image diffusion process \cite{ho2020denoising} to train our diffusion model. 
In variant \uppercase\expandafter{\romannumeral2} (\textit{HOI-IDiff w/o initializing noisy HOI image}), we use our \textit{HOI image} diffusion process. Yet, we start denoising from completely random noise instead of from the noisy \textit{HOI image}. 
Moreover, in variant \uppercase\expandafter{\romannumeral3} (\textit{HOI-IDiff with noisy HOI image as condition}), we still start denoising from completely random noise, but we use the noisy \textit{HOI image} as the condition for the reverse process. 
Lastly, in variant \uppercase\expandafter{\romannumeral4} (\textit{HOI-IDiff diffusion process w/o maintaining to be distributions}), instead of $1 - \beta_k$ and $\beta_k$ in Eq.~\ref{eq:forward_1}, we use back $\sqrt{1 - \beta_k}$ and $\sqrt{\beta_k}$ in Eq.~\ref{eq:revisiting_1} as regulators, and thus no longer keep the intermediate steps during the forward process to be a set of discrete probability distributions. 
As shown in Tab.~\ref{Tab:ablation_study_1}, compared to our framework, the performance of variant \uppercase\expandafter{\romannumeral1} drops significantly. This shows the importance of ensuring the compatibility between the image diffusion model and the \textit{HOI images}. Moreover, our framework also outperforms variants \uppercase\expandafter{\romannumeral2}-\uppercase\expandafter{\romannumeral4}. This further shows the efficacy of the different designs we made in our \textit{HOI image} diffusion process.

Next, we further build the diffusion model in our framework with a slice patchification architecture. To validate this design, we test other four variants. In variant \uppercase\expandafter{\romannumeral5} (\textit{HOI-IDiff with architecture in \cite{ho2020denoising}}), we use the typical diffusion model architecture in \cite{ho2020denoising} to build our model, while in variant \uppercase\expandafter{\romannumeral6} (\textit{HOI-IDiff with architecture in \cite{peebles2023scalable}}), we adopt the architecture in \cite{peebles2023scalable}. 
Moreover, in variant \uppercase\expandafter{\romannumeral7} (\textit{HOI-IDiff w/o horizontal slice patches}), we still employ the model architecture with the slice patchification operation but we patchify the \textit{HOI image} only into $W$ vertical slice patches.
Lastly, in variant \uppercase\expandafter{\romannumeral8} (\textit{HOI-IDiff w/o vertical slice patches}), we patchify the \textit{HOI image} only into $H$ horizontal slice patches.
As shown in Tab.~\ref{Tab:ablation_study_1}, our framework leveraging the slice patchfication model architecture consistently outperforms these four variants from variant \uppercase\expandafter{\romannumeral5} to variant \uppercase\expandafter{\romannumeral8}. This further shows the advantage of the slice patchification architecture in \textit{HOI image} generation.

\begin{table}[h]
\centering
\resizebox{\columnwidth}{!}
{
\small
\begin{tabular}{l|ccc}
\hline
Method & Full & Rare & Non-rare\\
\hline
\uppercase\expandafter{\romannumeral1}: $W \times 2$ \textit{HOI image} & 46.43 & 47.22 & 46.19 \\
\uppercase\expandafter{\romannumeral2}: $H$ \& $W \times 2$ \textit{HOI images} & 46.83 & 47.47 & 46.64 \\
\uppercase\expandafter{\romannumeral3}: $H \times W \times 2$ \textit{HOI image} &  47.71 & 48.36 & 47.52 \\
\uppercase\expandafter{\romannumeral4}: Box coordinates \& $H \times W \times 2$ \textit{HOI image} & 47.79 & 48.36 & 47.62 \\
\hline
\end{tabular}}
\vspace{-0.2cm}
\caption{Evaluation on the HOI image formulation process. Note that, even when we only generate $W \times 2$ \textit{HOI images}, we still achieve SOTA performance.}
\label{Tab:ablation_study_3}
\end{table}

\noindent\textbf{Impact of the HOI image formulation process.} In our framework, we formulate the \textit{HOI image} $I^{hoi}$ as the product of $v^{obj}$ of size $H$ and $m^{int}$ of size $W \times 2$ (\textit{$H \times W \times 2$ HOI image}).
To validate the efficacy of this formulation, we test three variants. 
In variant \uppercase\expandafter{\romannumeral1} (\textit{$W \times 2$ HOI image}), for each human-object pair, we first obtain its object category directly from the off-the-shelf object detector. Next, we generate only $m^{int}$, as the \textit{HOI image}, through the diffusion process. 
In variant \uppercase\expandafter{\romannumeral2} (\textit{$H$ \& $W \times 2$ HOI images}), we formulate $v^{obj}$ and $m^{int}$ as two separate \textit{HOI images}, and generate each of them through a separate diffusion process. 
Lastly, in variant \uppercase\expandafter{\romannumeral4} (\textit{Box coordinates \& $H \times W \times 2$ HOI image}), during diffusion, besides generating high-quality \textit{HOI images} of size $H \times W \times 2$, we also refine the bounding box coordinates for each human-object pair. As shown in Tab.~\ref{Tab:ablation_study_3}, our framework and these three variants all achieve better performances than the existing state-of-the-art method \cite{wu2024exploring}. Moreover, our framework (\textit{$H \times W \times 2$ HOI image}) outperforms variants \uppercase\expandafter{\romannumeral1} and \uppercase\expandafter{\romannumeral2}. Meanwhile, our framework achieves compatible performance compared to variant \uppercase\expandafter{\romannumeral4} which additionally refines bounding box coordinates. Thus, considering the framework's complexity as well, in our diffusion process, we generate $H \times W \times 2$ \textit{HOI images}, but do not additionally refine the bounding box coordinates.

\begin{figure}[h]
\centering
\includegraphics[width=\linewidth]{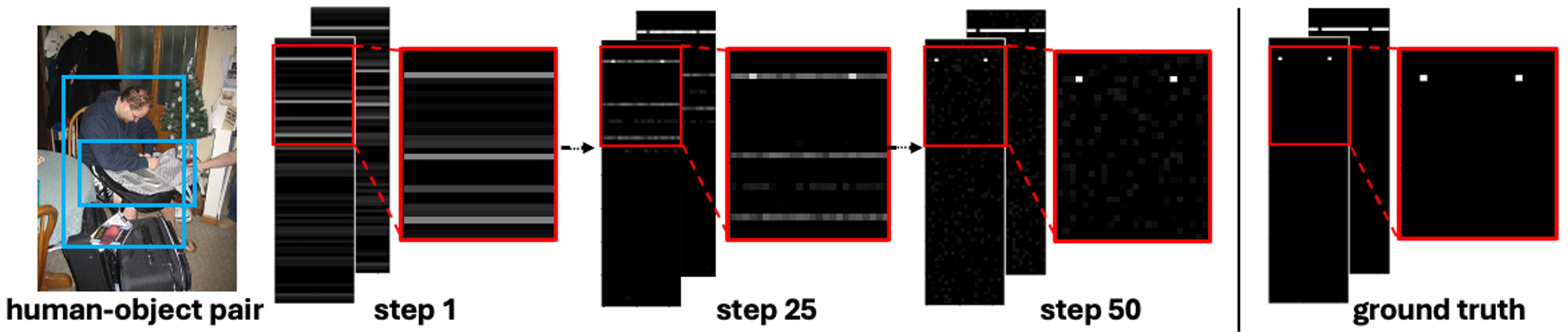}
\vspace{-0.5cm}
\caption{Visualization of the \textit{HOI image} diffusion process.}
\vspace{-0.2cm}
\label{fig:3}
\end{figure}
\noindent\textbf{Visualization of the diffusion process.} In Fig.~\ref{fig:3}, we visualize the \textit{HOI images} produced throughout the reverse diffusion process on V-COCO. As shown, our HOI-IDiff framework effectively eliminates noise and indeterminacy from the initially noisy \textit{HOI image} in a step-wise manner, ultimately generating a high-quality \textit{HOI image}. 
{For better visibility, the figure is normalized with log function.}

\section{Conclusion}
\label{sec:conclusion}

In this paper, from a novel perspective, we recast HOI detection as an \textit{HOI image} generation problem. 
Correspondingly, we propose an innovative image-diffusion-based HOI detection framework, which tackles the HOI detection task via generating high quality \textit{HOI images} through a carefully-designed \textit{HOI image} diffusion process. 
Our framework consistently achieves state-of-the-art performance on two commonly-used evaluation benchmarks.
{
    \small
    \bibliographystyle{ieeenat_fullname}
    \bibliography{main}
}


\end{document}